\documentclass[11pt,a4paper]{article}
\usepackage[hyperref]{emnlp2020}
\usepackage{times}
\usepackage{latexsym}

\usepackage[ruled,vlined]{algorithm2e}

\usepackage{amsmath}
\usepackage{cleveref}       
\usepackage{url}            
\usepackage{booktabs}       
\usepackage{mathtools}
\usepackage{amssymb}
\usepackage{bm}
\usepackage{bbm}
\usepackage{multirow}
\usepackage{amsfonts}       
\usepackage{nicefrac}       
\usepackage{microtype}

\newcommand{\x}{\bm{\mathrm{x}}}
\newcommand{\y}{\bm{\mathrm{y}}}
\newcommand{\yhat}{\hat{\y}}
\newcommand{\z}{\bm{\mathrm{z}}}
\newcommand{\MU}{\bm{\mathrm{\mu}}}

\newcommand{\dec}{p_\theta (\y|\z,\x)}
\newcommand{\pri}{p_\theta (\z|\x)}
\newcommand{\joint}{p_\theta (\y,\z|\x)}
\newcommand{\post}{q_\phi (\z|\y,\x)}

\DeclareMathOperator*{\E}{\mathbb{E}}
\newcommand{\e}{E_{\psi}(\z; \x)}
\newcommand{\score}{S_{\psi}(\z; \x)}
\newcommand{\gradz}{\nabla_{\z}}
\newcommand{\tauz}{\tau(\z)}
\newcommand{\zini}{\z}
\newcommand{\zfin}{\overline\z}
\newcommand{\zd}{\widetilde\z}
\newcommand{\TAU}{\tau_{\theta}(\z; \x)}

\newcommand{\wmtende}{WMT'14 En$\rightarrow$De}

\newcommand{\wmtroen}{WMT'16 Ro$\rightarrow$En}

\newcommand{\iwsltdeen}{IWSLT'16 De$\rightarrow$En}
\newcommand{\argmax}{\text{argmax}}
\newcommand{\argmin}{\text{argmin}}

\newcommand{\std}[1]{{\scriptsize$\pm {#1}$}}

\definecolor{pinegreen}{rgb}{0.0, 0.47, 0.44}
\definecolor{olive}{rgb}{0.5, 0.5, 0.0}
\definecolor{ao}{rgb}{0.0, 0.5, 0.0}
\definecolor{darkpastelgreen}{rgb}{0.01, 0.75, 0.24}
\definecolor{forestgreen}{rgb}{0.13, 0.55, 0.13}
\definecolor{htmlgreen}{rgb}{0.0, 0.5, 0.0}

\crefformat{section}{\S#2#1#3} 
\crefformat{subsection}{\S#2#1#3}
\crefformat{subsubsection}{\S#2#1#3}

\aclfinalcopy 
\def\aclpaperid{1432} 

\title{Iterative Refinement in the Continuous Space \\ for Non-Autoregressive Neural Machine Translation}

\author{Jason Lee \\
  New York University \\
  {\small \texttt{jason@cs.nyu.edu} } \\\And
  Raphael Shu \\
  \\
  {\small \texttt{zomux2000@gmail.com} } \\\And
  Kyunghyun Cho \\
  New York University \\
  {\small \texttt{kyunghyun.cho@nyu.edu} } \\
}

\date{}

\begin{document}
\maketitle

\begin{abstract}
We propose an efficient inference procedure for non-autoregressive machine translation that iteratively refines translation purely in the continuous space.
Given a continuous latent variable model for machine translation~\citep{shu20latent}, we train an inference network to approximate the gradient of the marginal log probability of the target sentence, using only the latent variable as input.
This allows us to use gradient-based optimization to find the target sentence at inference time that approximately maximizes its marginal probability.
As each refinement step only involves computation in the latent space of low dimensionality (we use 8 in our experiments), we avoid computational overhead incurred by existing non-autoregressive inference procedures that often refine in token space.
We compare our approach to a recently proposed EM-like inference procedure~\citep{shu20latent} that optimizes in a hybrid space, consisting of both discrete and continuous variables.
We evaluate our approach on {\wmtende}, {\wmtroen} and {\iwsltdeen}, and observe two advantages over the EM-like inference:
(1) it is computationally efficient, i.e. each refinement step is twice as fast,
and (2) it is more effective, resulting in higher marginal probabilities and BLEU scores with the same number of refinement steps.
On {\wmtende}, for instance, our approach is able to decode $6.2$ times faster than the autoregressive model with minimal degradation to translation quality (0.9 BLEU).
\end{abstract}

\section{Introduction}
\label{intro}

Most neural machine translation systems are autoregressive, hence decoding latency grows linearly with respect to the length of the target sentence. For faster generation, several work proposed non-autoregressive models with sub-linear decoding latency given sufficient parallel computation~\citep{gu18non,lee18deterministic,kaiser18fast}. 

As it is challenging to precisely model the dependencies among the tokens without autoregression,
many existing non-autoregressive models first generate an initial translation which is then iteratively refined to yield better output~\citep{lee18deterministic,gu19insertion,gha19mask}. 
While various training objectives are used to admit refinement (e.g. denoising, evidence lowerbound maximization and mask language modeling), the generation process of these models is similar in that the refinement process happens in the \emph{discrete} space of sentences.

Meanwhile, another line of work proposed to use \emph{continuous} latent variables for non-autoregressive translation, such that the distribution of the target sentences can be factorized over time given the latent variables~\citep{ma19flowseq,shu20latent}. 
Unlike the models discussed above, finding the most likely target sentence under these models requires searching over continuous latent variables.
To this end, \citet{shu20latent} proposed an EM-like inference procedure that optimizes over a hybrid space consisting of both continuous and discrete variables.
By introducing a deterministic delta posterior, it maximizes a proxy lowerbound by alternating between matching the delta posterior to the original approximate posterior (continuous optimization), and finding a target sentence that maximizes the proxy lowerbound (discrete search).

In this work, we propose an iterative inference procedure for latent variable non-autoregressive models that purely operates in the continuous space.\footnote{We open source our code at {\texttt{\url{https://github.com/zomux/lanmt-ebm}}}}
Given a latent variable model, we train an inference network to estimate the gradient of the marginal log probability of the target sentence, using only the latent variable as input.
At inference time, we find the target sentence that approximately maximizes the log probability by
(1) initializing the latent variable e.g. as the mean of the prior, and
(2) following the gradients estimated by the inference network.

We compare the proposed approach with the EM-like inference~\citep{shu20latent} on three machine translation datasets: {\wmtende}, {\wmtroen} and {\iwsltdeen}.
The advantages of our approach are twofold:
(1) each refinement step is twice as fast, as it avoids discrete search over a large vocabulary, and 
(2) it is more effective, giving higher marginal probabilities and BLEU scores with the same number of refinement steps.
Our procedure results in significantly faster inference, for instance giving $6.2\times$ speedup over the autoregressive baseline on {\wmtende} at the expense of $0.9$ BLEU score.

\section{Background: Iterative Refinement for Non-Autoregressive Translation}
\label{sec:background}

We motivate our approach by reviewing existing refinement-based non-autoregressive models for machine translation in terms of their inference procedure.
Let us use $V,\:D,\:T$ and $L$ to denote vocabulary size, latent dimensionality, target sentence length and the number of refinement steps, respectively.

Most machine translation models are trained to maximize the conditional log probability $\log p(\y|\x)$ of the target sentence $\y$ given the source sentence $\x$, averaged over the training data consisting of sentence pairs $\{(\x_n, \y_n)\}_{n=1}^N$.
To find the most likely target sentence at test time, one performs maximum-a-posteriori inference by solving a search problem $\hat{\y} = \argmax_{\y} \log p(\y|\x).$ 

\subsection{Refinement in a Discrete Space}

As the lack of autoregression makes it challenging to model the dependencies among the target tokens, most of the existing non-autoregressive translation models use iterative refinement to impose dependencies in the generation process.
Various training objectives are used to incorporate refinement, e.g. denoising~\citep{lee18deterministic}, mask language modeling~\citep{gha19mask} and evidence lowerbound maximization~\citep{chan19kermit,gu19insertion}.
However, inference procedures employed by these models are similar in that an initial hypothesis is generated and then successively refined.
We refer the readers to \citep{mansimov19generalized} for a formal definition of a sequence generation framework that unifies these models, and briefly discuss the inference procedure below.

By viewing each refinement step as introducing a discrete random variable $\z_i$ (a $T \times V$-dimensional matrix, where each row is one-hot), 
inference with $L$ refinement steps requires finding $\y$ that maximizes the log probability $\log p(\y|\x)$.
\begin{align}
\notag
&\log p_{\theta}(\y|\x) = \log \sum_{\z_{1:L}} p_\theta(\y,\z_{1:L}|\x) \\
\notag
&= \log \sum_{\z_{1:L}} \Big( p_\theta(\y | \z_{1:L}, \x) \cdot \prod_{i=1}^{L} p_\theta(\z_{i}|\z_{<i}, \x) \Big) \\
&\geq \sum_{\z_{1:L}} \Big( \log p_\theta(\y | \z_{1:L}, \x) + \sum_{i=1}^{L} \log p_\theta(\z_{i}|\z_{<i}, \x) \Big).
\end{align}

\noindent As the marginalization over $\z_{1:L}$ is intractable, inference for these models instead maximize the log joint probability with respect to $\hat{\z}_{1:L}$ and $\y$:
\begin{align*}
\log p_\theta(\y | \hat{\z}_{1:L}, \x) + \,\, \sum_{i=1}^{L} \log p_\theta(\hat{\z}_{i}|\hat{\z}_{<i}, \x).
\end{align*}

\noindent Approximate search methods are used to find $\hat{\z}_{1:L}$ as $\hat{\z}_{i} = \argmax_{\z_i} \log p_\theta(\z_i|\hat{\z}_{<i},\x).$


\subsection{Refinement in a Hybrid Space}
\label{sec:lvnar}

\paragraph{Learning}
On the other hand, \citet{ma19flowseq,shu20latent} proposed to use \emph{continuous} latent variables for non-autoregressive translation.
By letting the latent variables $\z$ (of dimensionality $T \times D$) capture the dependencies between the target tokens, the decoder $\dec$ can be factorized over time.
As exact posterior inference and learning is intractable for most deep parameterized prior and decoder distributions, these models are trained to maximize the evidence lowerbound (ELBO)~\citep{kingma14auto,wainwright08graphical}.
\begin{align*}
    \log p_\theta(\y|\x) \geq \E_{\z \sim q_\phi} \Big[ \log \frac{\joint}{\post} \Big]
\end{align*}

\paragraph{Inference}
Exact maximization of ELBO with respect to $\y$ is challenging due to the expectation over $\z \sim q_\phi$.
To approximately maximize the ELBO, \citet{shu20latent} proposed to optimize a deterministic proxy lowerbound using a Dirac delta posterior: 
\begin{equation*}
\delta(\z|\MU) = \mathbbm{1}_{\MU}(\z)
\end{equation*}

\noindent Then, the ELBO reduces to the following proxy lowerbound:
\vskip -0.2in
\begin{align*}
\notag
& \E_{\z\sim \delta(\z|\MU)}{\Big[ \dec + \pri \Big]} + \overbrace{\mathcal{H}(\delta)}^{=0}, \notag \\
&= \log p_{\theta} (\y|\MU, \x) + \log p_{\theta} (\MU|\x).
\end{align*}

\noindent \citet{shu20latent} proposed to approximately maximize the ELBO with an EM-like inference procedure, to which we refer as \emph{delta inference}. It alternates between continuous and discrete optimization:
(1) E-step matches the delta posterior with the approximate posterior by minimizing their KL divergence: \mbox{$\MU_i = \argmin_{\MU}\text{KL}\big[ \,\delta(\z|\MU) \,\big\Vert\, q_\phi(\z | \hat{\y}_{i-1}, \x) \,\big]$}, and
(2) M-step maximizes the proxy lowerbound with respect to $\y$: $\hat{\y}_i = \argmax_{\y} \log p_\theta (\y|\MU_i, \x).$ 
Overall, delta inference finds $\y$ and $\MU$ that maximizes $\log p_\theta(\y|\MU, \x) + \log q_\phi(\MU|\y, \x).$
This iterative inference procedure in hybrid space was empirically shown to result in improved BLEU scores and ELBO on each refinement step~\citep{shu20latent}.

\section{Iterative Refinement in a Continuous Space}
\label{model}
While the delta inference procedure is an effective inference algorithm for machine translation models with continuous latent variables, it is unsatisfactory as the M-step requires searching over $V$ tokens $T$ times for each refinement step. 
As $V$ is large for most machine translation models, this is an expensive operation, even when the $T$ searches can be parallelized.
We thus propose to replace the delta inference with continuous optimization in the latent space only, given the underlying latent variable model.

\subsection{Learning}

Let us define $\TAU$ as the marginal log probability of the most likely target sentence under the latent variable model given $\z$.
\begin{align}
\label{eq:ebm}
    &\TAU = \log p_\theta(\hat{\y}|\x),
\end{align}

\noindent where $\yhat = \argmax_{\y} \log p_{\theta}(\y|\z,\x).$ Our goal is to find a function $-E_\psi(\z; \x)$ that approximates $\TAU$ up to an additive constant and a positive multiplicative factor, such that
\begin{align}
\notag
    &\argmin_{\z} \big(E_\psi(\z; \x)\big) \approx \argmax_{\z} \big(\TAU\big).
\end{align}

\noindent In this work, instead of directly approximating $\tau_{\theta}$, we train $-E_\psi$ to learn the difference of $\tau_{\theta}$ 
between a pair of configurations of latent variables.
Omitting the source sentence $\x$ and the model parameters $\theta$ for notational simplicity, we solve the following problem for $\zini \neq \zfin$:
\begin{align}
\label{eq:ebmmid}
\notag
\min_\psi\: &\Big\Vert \Big(-E_\psi(\zfin) + E_\psi(\zini)\Big) -\Big(\tau(\zfin) - \tau(\zini)\Big) \Big\Vert^2 \\
\approx \min_\psi\: &\Big\Vert \gradz{E_\psi(\z)} \Big\Vert^2 + 2\,\Big( \big( \gradz{E_\psi(\z)} \big)^\intercal \cdot \gradz\tauz \Big).
\end{align}

\noindent See Appendix~\ref{sec:app} for a full derivation.
Intuitively, $\gradz\big(-\e\big)$ is trained to approximate $\gradz\,\TAU$, as Eq.~\ref{eq:ebmmid} maximizes their dot product while minimizing its squared norm.

As $\TAU$ is not differentiable with respect to $\z$ due to the argmax operation in Eq.~\ref{eq:ebm}, $\gradz\,\TAU$ is not defined.
We thus use a proxy gradient from delta inference.
Furthermore, we weigh the latent configuration $\z$ according to the prior.
Our final training objective for $E_\psi$ is then as follows:
\begin{align}
\notag
\mathbb{E}_{\z \sim p_\theta(\z|\x)} \bigg [ \, & \Big\Vert \gradz\e \Big\Vert^2 \, +\\
\label{eq:ebmfinal}
 & 2 \, \Big( \big( \gradz {\e} \big)^\intercal \cdot (\zd - \z) \Big) \bigg],
\end{align}
where $\zd$ is the output of applying $k$ steps of delta inference on $\z$.
If delta inference improves the log probability at each iteration, we hypothesize that $(\zd - \z)$ is a reasonable approximation to the true gradient $\gradz\,\TAU$.
We empirically show that this is indeed the case in Sec.~\ref{sec:ll}.

\subsection{Parameterization}
We have two options for parameterizing $\gradz\e$ when minimizing Eq.~\ref{eq:ebmfinal}.
First, we can parameterize it as the gradient of a scalar-valued function $E$, to which earlier work have referred as an \emph{energy} function~\citep{teh03energy,lecun06tutorial}.
Second, we can parameterize it as a function $\score$ that directly outputs the gradient of the log probability with respect to $\z$ (which is often referred to as a \emph{score} function~\citep{hyvarinen05estimation}), without estimating the energy directly. 

While previous work found direct score estimation that bypasses energy estimation unstable~\citep{alain14what,saremi18deep}, it leads to faster inference by avoiding backpropagation in each refinement step.
We compare the two approaches in our experiments.

\subsection{Inference}

At inference time, we initialize the latent variable (e.g. using either a sample from the prior or its mean) 
and iteratively update the latent variable using the estimated gradients (see Alg.~\ref{alg:ebm}).
As our inference procedure only involves optimization in the continuous space each step,
we avoid having to search over a large vocabulary.
We can either perform iterative refinement for a fixed number of steps, or until some convergence condition is satisfied.

\begin{algorithm}[t]
\SetAlgoLined
\SetKwInOut{Input}{Input}\SetKwInOut{Output}{Output}
\Input{\:$\x$, $\alpha$, $\theta$, $\psi$}
\Output{\:$\hat{\y}$}
 let $\z = \E_{\z \sim p_\theta(\z|\x)}[\z]$\\
 \While{termination condition not met,}{
    $\z = \z - \alpha \cdot ({\nabla_{\z} E_\psi(\z;\x)})$ \\
 }
 $\hat{\y} = \argmax_{\y} \log p_\theta(\y| \z, \x)$
 \caption{Inference for Latent Variable Models using Learned Gradients}
 \label{alg:ebm}
\end{algorithm}

\begin{table*}[!t]
\small
\centering
\begin{sc}
\begin{tabular}{llp{1.0cm}rrrrrrrrrrr} \toprule
& & & \multicolumn{3}{c}{\wmtende} & & \multicolumn{3}{c}{\wmtroen} & & \multicolumn{3}{c}{\iwsltdeen}  \\ 
& & & \multicolumn{1}{c}{Bleu} & \multicolumn{1}{c}{Speed} & \multicolumn{1}{c}{ Time} & & \multicolumn{1}{c}{Bleu} & \multicolumn{1}{c}{Speed} & \multicolumn{1}{c}{ Time} & & \multicolumn{1}{c}{Bleu} & \multicolumn{1}{c}{Speed} & \multicolumn{1}{c}{ Time} \\ 
 
\midrule
\multicolumn{2}{c}{ \multirow{2}{*}{ \rotatebox{90}{\parbox{0.70cm}{\centering AR}} } }
& $b=1$ & 27.5  & $1.1\times$  & 251 \std{175} &  & 30.9 & $1.1\times$ & 511 \std{560} &  & 31.1 & $1.1 \times$ & 178 \std{139} \\
\multicolumn{2}{c}{  }
& $b=4$ & 28.3  &  $1\times$  & 291 \std{194} & & 31.5 & $1 \times$ & 610 \std{630} & & 31.5 & $1 \times$ & 210 \std{161}  \\
    
\midrule

\multirow{15}{*}{\rotatebox{90}{\parbox{5.5cm}{\centering NAR LVM}}}

& \multirow{5}{*}{\rotatebox{90}{\parbox{1.4cm}{\centering Delta}}} 
  & $L=0$ & 25.7 & $15\times$ & 19 \std{1} &  & 28.4 & $34 \times$  & 18 \std{5}  & {} & 27.0 & $19 \times$ & 11 \std{5} \\
& & $L=1$ & 26.1  & $6.3\times$ & 46 \std{5} &  & 29.0 & $19 \times$ & 32 \std{5}   & {} & 28.3 & $11 \times$ & 18 \std{6}  \\
& & $L=2$ & 26.2 & $4.0\times$ & 72 \std{3}  &  & 29.1 & $14 \times$  & 45 \std{7}  & {} & 28.5 & $8.0 \times$ & 26 \std{7}  \\
& & $L=4$ & 26.1 & $2.8\times$ & 103 \std{5} &  & 29.1 & $8.5 \times$  & 72 \std{5}   & {} & 28.6 & $5.2 \times$ & 40 \std{9}  \\
& & Search & 26.9 & $5.5\times$ & 63 \std{8} & & 30.3 & $13 \times$ & 48 \std{7} &       & 29.7  & $6.0 \times$  & 35 \std{6}  \\

\cmidrule{2-14}
& \multirow{5}{*}{\rotatebox{90}{\parbox{1.6cm}{\centering Energy}}}
  & $L=0$ & 25.7 & 15$\times$ & 19 \std{1} & {} & 28.4  & $34 \times$ & 18 \std{5}  & {} & 27.0 & $19 \times$ & 11 \std{5}  \\
& & $L=1$ & 26.1 & 5.8$\times$ & 50 \std{3}  & {} & 28.8 & $17 \times$ & 36 \std{5}  & {} & 28.6 & $9.5 \times$ & 22 \std{7}   \\
& & $L=2$ & 26.1 & 4.2$\times$  & 69 \std{4} & {} & 28.9 & $11 \times$ & 55 \std{9}  & {} & 28.7 & $7.0 \times$& 30 \std{9}  \\
& & $L=4$ & 26.0 & 2.5$\times$  & 117 \std{6} & {} & 28.8 & $7.1 \times$ & 85 \std{5} & {} & 28.8 & $4.5 \times$ & 46 \std{9}  \\
& & Search & 27.1 & 4.4$\times$  & 66 \std{9} & {} & 30.4 & $12 \times$ & 53 \std{7} & {} & 29.9  & $5.0 \times$ & 42 \std{7}   \\

\cmidrule{2-14}
& \multirow{5}{*}{\rotatebox{90}{\parbox{1.6cm}{\centering Score}}} 
  & $L=0$ & 25.7 & $15\times$ & 19 \std{1} &  & 28.4 & $34 \times$  & 18 \std{5}  & {} & 27.0 & $19 \times$ & 11 \std{5} \\
& & $L=1$ & 26.3 & $10\times$ & 29 \std{2} &  & 29.1 & $24 \times$  & 25 \std{5}  & {} & 28.8 & $13 \times$ & 16 \std{6} \\
& & $L=2$ & 26.3  & $7.6\times$ & 38 \std{2} &  & 29.1 & $19 \times$  & 32 \std{6}  & {} & 29.0 & $10 \times$ & 20 \std{5} \\
& & $L=4$ & 26.3 & $5.7\times$ & 51 \std{4} &  & 29.1 & $14 \times$  & 44 \std{5} & {} & 29.1 & $7.5 \times$ & 28 \std{5} \\
& & Search & \textbf{27.4} & $\bm{6.2}\times$ & 47 \std{8} & {} & \textbf{30.4}    &  $\bm{15} \times$  &  41 \std{6}    & {} &   \textbf{30.2}    & $\bm{6.3} \times$ & 33 \std{4}   \\
\bottomrule
\end{tabular}
\end{sc}
\caption{Translation quality and inference speed of autoregressive baseline (AR) and several inference procedures for non-autoregressive latent variable model (NAR LVM): Delta inference (\emph{Delta})~\citep{shu20latent}, the proposed inference procedure with estimated energy (\emph{Energy}) or score (\emph{Score}).
\emph{Speed}: inference speedup compared to the autoregressive model with beam width 4.
\emph{Time}: Average wall clock time per example in milliseconds on a Tesla V100 GPU (with standard deviations).
$b$: beam width, $L$: the number of refinement steps.
\emph{Search}: parallel decoding with $5$ length candidates and $5$ samples from the prior, with $1$ refinement step. 
Results above \emph{Search} are obtained by initializing the latent variable as the mean of the prior.
We boldface the highest BLEU among the latent variable models.
}
\label{tab:mainresult}
\end{table*}

\section{Experimental Setup}
\label{exp}

\subsection{Datasets and Preprocessing}
We evaluate our approach on three widely used machine translation datasets: {\iwsltdeen}\footnote{\url{https://wit3.fbk.eu/}} (containing 197K training, 2K development and 2K test sentence pairs), {\wmtroen}\footnote{\url{www.statmt.org/wmt16/translation-task.html}} (612K, 2K, 2K pairs) and {\wmtende}\footnote{\url{www.statmt.org/wmt14/translation-task.html}} (4.5M, 3K, 3K pairs).

We use sentencepiece tokenization~\citep{kudo18sentencepiece} with 32K sentencepieces on all datasets. For {\wmtroen}, we follow \citet{sennrich16edinburgh} and normalize Romanian and remove diacritics before applying tokenization. For training, we discard sentence pairs if either the source or the target length exceeds 64 tokens. 

Following \citet{lee18deterministic}, we remove repetitions from the translations with a simple postprocessing step before computing BLEU scores.
We use detokenized BLEU with Sacrebleu~\citep{post18call}.

\paragraph{Distillation}
Following previous work on non-autoregressive translation, we train non-autoregressive models on the target sentences generated by an autoregressive model~\citep{kim16sequence,gu18non} trained using the FairSeq framework~\citep{ott19fairseq}.

\subsection{Models and Baselines}
\label{sec:exp:model}
\paragraph{Autoregressive baselines}
We use Transformers~\citep{vaswani17attention} with the following hyperparameters.
For {\wmtroen} and {\wmtende}, we use Transformer-base.
For {\iwsltdeen}, we use a smaller model with $(d_\text{model}, \,d_\text{filter}, \,n_\text{layers}, \, n_\text{heads}) = (256, \:1024, \:5, \:2)$.

\paragraph{Non-autoregressive latent variable models}
We closely follow the implementation details from \citep{shu20latent}.
The prior and the approximate posterior distributions are spherical Gaussian distributions with learned mean and variance, and the decoder is factorized over time.
The only difference is at inference time, the target sentence length is predicted once and fixed throughout the refinement procedure.
Therefore, the latent variable dimensionality $\mathbb{R}^{T \times D}$ does not change.

The decoder, prior and approximate posterior distributions are all parameterized using $n_\text{layers}$ Transformer decoder layers (the last two also have a final linear layer that outputs mean and variance). 
For {\iwsltdeen}, we use $(d_\text{model}, \,d_\text{filter}, \,n_\text{layers}, \, n_\text{heads}) = (256, \:1024, \:3, \:4)$. For {\wmtende} and {\wmtroen}, we use $(512, \:2048, \:6, \:8)$.
The latent dimensionality $d_\text{latent}$ is set to 8 across all datasets.
The source sentence encoder is implemented with a standard Transformer encoder. Given the hidden states of the source sentence, the length predictor (a 2-layer MLP) predicts the length difference between the source and target sentences as a categorical distribution in $[-50, 50].$

\paragraph{Energy function}
$E_\psi(\z; \x)$ is parameterized with $n_\text{layers}$ Transformer decoder layers and a final linear layer with the output dimensionality of 1.
We average the last Transformer hidden states across time and feed it to a linear layer to yield a scalar energy value.

\paragraph{Score function}
When directly estimating the gradient of the log probability with respect to $\z$, $S_\psi(\z; \x)$ is parameterized with $n_\text{layers}$ Transformer decoder layers and a final linear layer with the output dimensionality of $d_\text{latent}.$

\subsection{Training and Optimization}
We use the Adam optimizer~\citep{kingma15adam} with batch size of 8192 tokens and the learning rate schedule used by \citet{vaswani17attention} with warmup of 8K steps. 
When training our inference networks, we fix the underlying latent variable model.
Our inference networks are trained for 1M steps to minimize Eq.~\ref{eq:ebmfinal}, where $\zd$ is obtained by applying $k(=4)$ iterations of delta inference on $\z$ sampled from the prior.
We also find that stochastically applying one gradient update (using the estimated gradients) to $\z$ before computing $\zd$ leads to better performance.

\subsection{Inference}

\paragraph{Step size}
For the proposed inference procedure, we use the step size $\alpha=1.0$ as it performed well on the development set. 


\paragraph{Length prediction}
Given a distribution of target sentence length, we can either
(1) take the argmax, or
(2) select the top $l$ candidates and decode them in parallel~\citep{gha19mask}.
In the second case, we select the output candidate with the highest log probability under an autoregressive model, normalized by its length.

\paragraph{Latent search}
In Alg.~\ref{alg:ebm}, we can either initialize the latent variable with a sample from the prior, or its mean.
We use $n_w$ samples from the prior and perform iterative refinement (e.g. delta inference or the proposed inference procedures) in parallel.
Similarly to length prediction, we select the output with the highest log probability.
To avoid stochasticity, we fix the random seed during sampling.

\section{Quantitative Results}

\subsection{Translation Quality and Speed}

Table~\ref{tab:mainresult} presents translation performance and inference speed of several inference procedures for the non-autoregressive latent variable models, along with the autoregressive baselines.
We emphasize that the same underlying latent variable model is used across three different inference procedures (Delta, Energy, Score), to compare their efficiency and effectiveness.

\paragraph{Translation quality}
We observe that both of the proposed inference procedures result in improvements in translation quality with more refinement steps.
For instance, 4 refinement steps using the learned score function improves BLEU by 2.1 on {\iwsltdeen}.
Among the proposed inference procedures, we find it more effective to use a learned score function, as it gives comparable or better performance to delta inference on all datasets.
A learned energy function results in comparable performance to delta inference.
Parallel decoding over multiple target length candidates and sampled latent variables leads to significant improvements in BLEU, resulting in 1 BLEU increase or more on all datasets.
Similarly to delta inference, we find that the proposed iterative inference procedures converge quite quickly, and often 1 refinement step gives comparable translation quality to running 4 refinement steps.

\begin{figure}[t]
\centering
  \includegraphics[width=.50\linewidth]{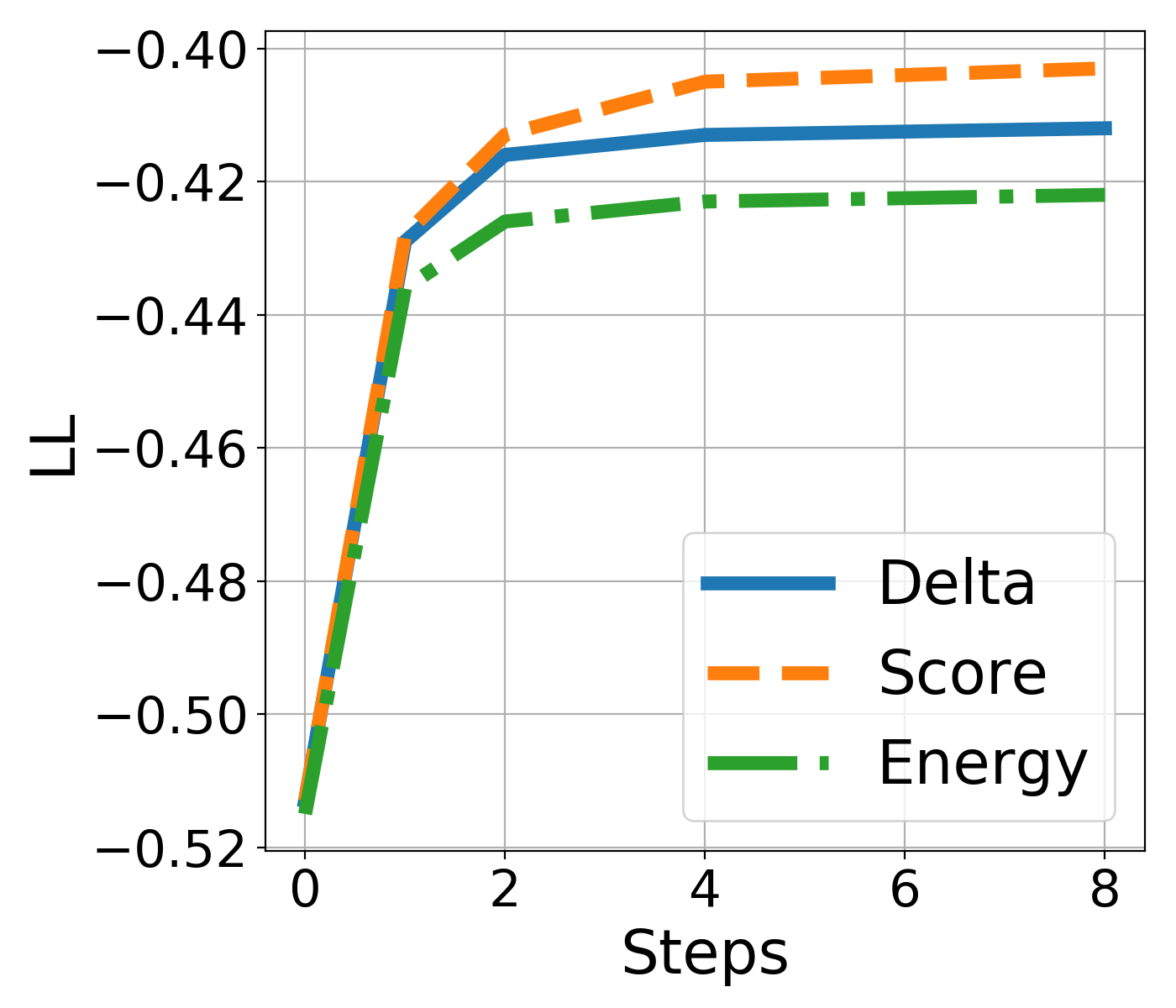}
\caption{Marginal log probability $\log p_\theta(\hat{\y}|\x)$ of output $\hat{\y}$ from each refinement step.}
\label{fig:elbo}
\end{figure}

\paragraph{Inference speed}
We observe that using a learned score function is significantly faster than delta inference: twice as fast on {\iwsltdeen} and {\wmtroen} and almost four times as fast on {\wmtende}.
On {\wmtende}, the decoding latency for 4 steps using the score is close to (within one standard deviation of) running 1 refinement step of delta inference.
On the other hand, we find that using the learned energy function is slower, presumably due to the overhead from backpropagation. We find its wall clock time to be similar to delta inference.
As the entire inference process can be parallelized, we find that parallel decoding with multiple length candidates and latent variable samples only incurs minimal overhead. 
Finally, we confirm that decoding latency for non-autoregressive models is indeed constant with respect to the sequence length (given parallel computation), as the standard deviation is small ($< 10$ ms) across test examples.

\paragraph{Overall result}
Overall, we find the proposed inference procedure using the learned score function highly effective and efficient. On {\wmtende}, using 1 refinement step and parallel search leads to $6.2 \times$ speedup over the autoregressive baseline with minimal degradation to translation quality (0.9 BLEU score).

\begin{figure}[t]
\centering
\begin{minipage}{.44\columnwidth}
  \centering
  \includegraphics[width=.98\linewidth]{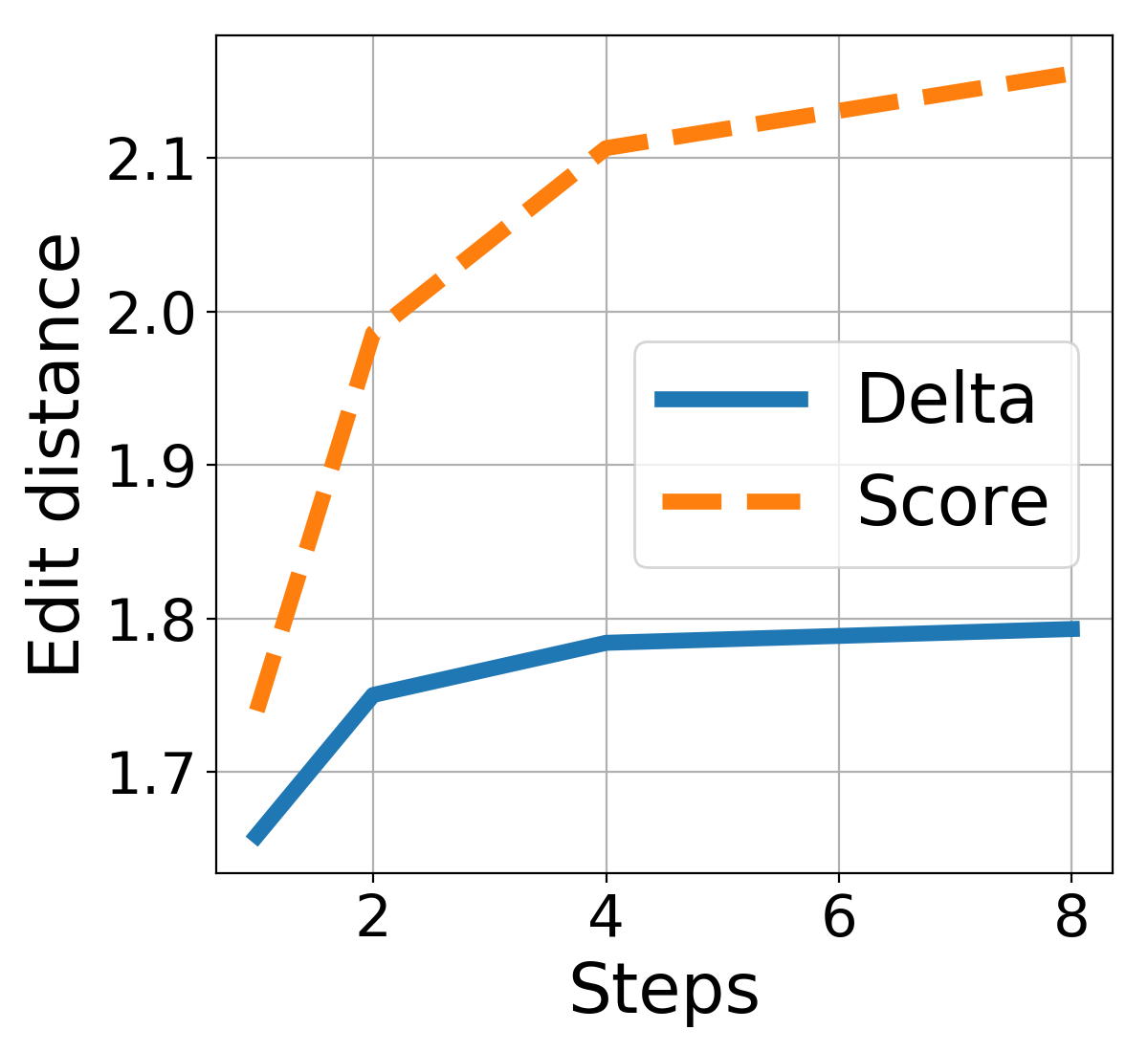}
\end{minipage}%
\begin{minipage}{.44\columnwidth}
  \centering
  \includegraphics[width=.98\linewidth]{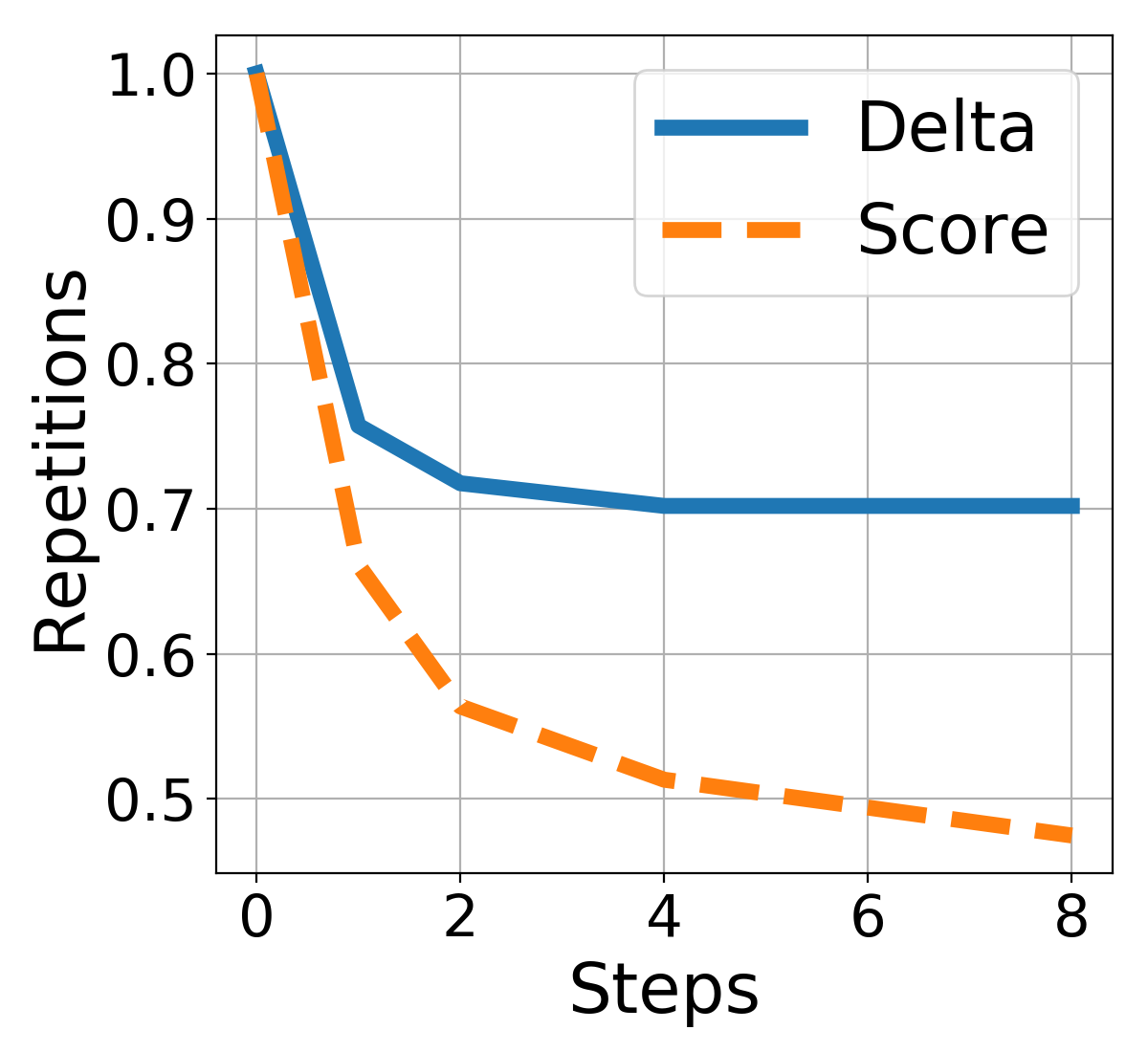}
\end{minipage}
\caption{Edit distance from the first output (left) and the number of repetitions in the output (right) for $L=\{1,2,4,8\}$ refinement steps for delta inference and inference using a learned score function.}
\label{fig:token}
\end{figure}

\begin{figure*}[!t]
\begin{center}
\centerline{\includegraphics[width=2.10\columnwidth]{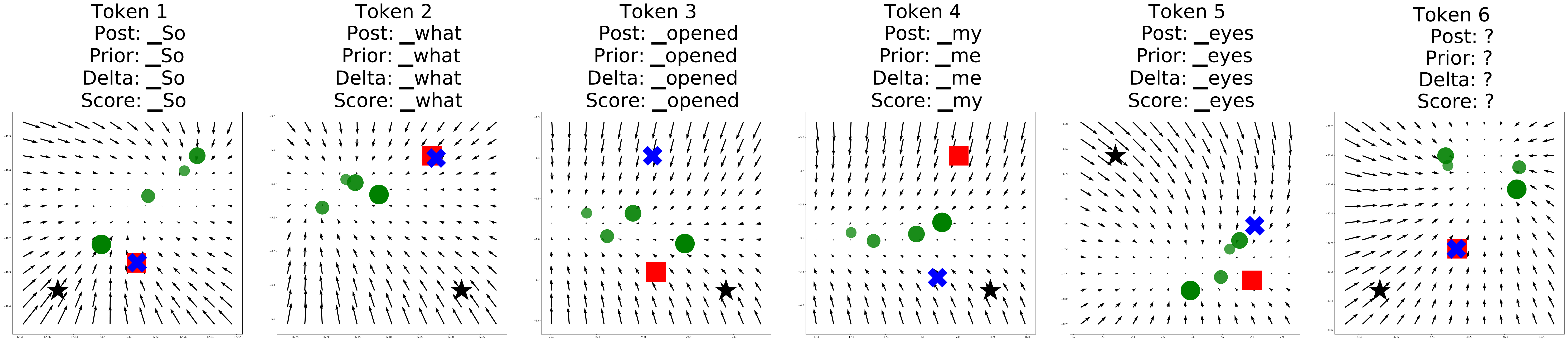}}
\caption{Visualization of estimated gradients and optimization trajectory.
Above each plot are tokens predicted from the following latent variables:
(1) approximate posterior mean,
(2) prior mean,
(3) delta inference and
(4) inference with the learned score.
{\textbf{Black star}}: latent variable before refinement (prior mean).
{\color{blue} \textbf{Blue cross}}: latent variables after $L = \{1, 2, 3, 4\}$ steps of delta inference (collapsed into a single point).
{\color{htmlgreen} \textbf{Green circle}}: latent variables after $L$ steps of inference with a learned score function. Marker size decreases with successive refinement steps.
{\color{red}\textbf{Red square}}: approximate posterior mean.
}
\label{fig:gradfield}
\end{center}
\vskip -0.20in
\end{figure*}

\subsection{Log Probability Comparison}
\label{sec:ll}

In Fig~\ref{fig:elbo}, we report the marginal log probability $\log p_{\theta}(\hat{\y}| \x)$ of $\hat{\y}$ found after $L$ steps of each iterative inference procedure on {\iwsltdeen}.
We estimate the marginal log probability by importance sampling with 500 samples from the approximate posterior.
We observe that the log probability improves with more refinement steps for all inference procedures (delta inference and the proposed procedures).
We draw two conclusions from this.
First, delta inference indeed increases log probability at each iteration.
Second, the proposed optimization scheme increases the target objective function it was trained on (log probability).

\subsection{Token Statistics}

We compare delta inference and the proposed inference with a learned score function in terms of token statistics in the output translations on {\iwsltdeen}.
In Figure~\ref{fig:token} (left), we compute the average edit distance (in sentencepieces) per test example from the initial output (mean of the prior).
It is clear that each refinement step using a learned score function results in more changes in terms of edit distance than delta inference.
In Figure~\ref{fig:token} (right), we compute the number of token repetitions in the output translations (before removing them in a post-processing step), relative to the initial output. We observe that refining with a learned score function results in less repetitive output compared to delta inference.

\section{Qualitative Results }

\subsection{Visualization of learned gradients}

We visualize the learned gradients and the optimization trajectory in Figure~\ref{fig:gradfield}, from a score inference network trained on a two-dimensional latent variable model on {\iwsltdeen}.
The example used to generate the visualization is shown below. 

\begin{table}[h!]
\small
\centering
\begin{tabular}{rl} \toprule
\multicolumn{1}{r|}{Source} & Was \"{o}ffnete mir also die Augen? \\
\multicolumn{1}{r|}{Reference} & So what opened {\color{red}my} eyes ? \\ \midrule
\multicolumn{1}{r|}{Posterior} & So what opened {\color{red}my} eyes ? \\
\multicolumn{1}{r|}{Prior} & So what opened me eyes ? \\
\multicolumn{1}{r|}{Delta} & So what opened me eyes ? \\
\multicolumn{1}{r|}{Score} & So what opened {\color{red}my} eyes ? \\
\bottomrule
\end{tabular}
\end{table}

\noindent 
We observe that for tokens 1, 2 and 6, delta inference converges quickly to the approximate posterior mean.
We also find that the local optima estimated by the score function do not necessarily coincide with the approximate posterior mean.
For Token 4, while the local optima estimated by the score function (green circle) is far from the posterior mean (red square), they both map to the reference translation (``my''), indicating that there exist multiple latent variables that map to the reference output.


\subsection{Sample translations}

We demonstrate that refining in the continuous space results in non-local, non-trivial revisions to the original sentence.
For each example in Table~\ref{tab:gen}, we show the English source sentence, German reference sentence, original translation decoded from a sample from the prior, and the revised translation with one gradient update using the estimated score function.

In Example 1, the positions of the main clause (``Es gibt nicht viele \"Arzte'') and the prepositional phrase (``im westafrikanischen Land'') are reversed in the continuous refinement process.
Inside the main clause, ``es gibt'' is revised to ``gibt es'', a correct grammatical form in German when the prepositional phrase comes before the main clause.

In Example 2, the two numbers are exchanged (`` 1,2 Milliarden Dollar'' and `` 6,9 Milliarden Dollar'') in the revised translation. Also, the phrase ``aus den'' (out of the) is correctly inserted between the two.

In Example 3, the noun phrase ``Weisheit in Bedouin'' is combined into a single German compound noun ``Bedouin-Weisheit''.
Also, the phrases ``Der erste ...'' and ``mit dieser ...'' are swapped in the refinement process, to better resemble the reference sentence.

\begin{table*}[!t]
\small
\centering
\begin{tabular}{rl}

\multicolumn{2}{l}{Example 1} \\
\toprule
\multicolumn{1}{r|}{Source} & There aren 't many doctors in the west African country ; just one for every 5,000 people \\

\multicolumn{1}{r|}{Reference} & In dem westafrikanischen Land gibt es nicht viele Ärzte, nur einen für 5.000 Menschen \\

\multicolumn{1}{r|}{Original} & {\color{blue}Es gibt nicht viele Ärzte} {\color{red}im westafrikanischen Land}, nur eine für 5.000 Menschen. \\

\multicolumn{1}{r|}{Refined} & {\color{red}Im westafrikanischen Land} {\color{blue}gibt es nicht viele Ärzte}, nur eine für 5.000 Menschen. \\

\\

\multicolumn{2}{l}{Example 2} \\
\toprule
\multicolumn{1}{r|}{Source} & Costumes are expected to account for \$ 1.2 billion dollars out of the \$ 6.9 billion spent , according to the NRF . \\

\multicolumn{1}{r|}{Reference} & Die Kostüme werden etwa 1,2 Milliarden der 6,9 Milliarden ausgegebenen US-Dollar ausmachen, so der NRF. \\

\multicolumn{1}{r|}{Original} & Es wird von, Kostüme, dass sie die dem NRF ausgegebenen {\color{blue}6,9 Milliarden Dollar} {\color{red}1,2 Milliarden Dollar} \\
\multicolumn{1}{r|}{} & ausmachen. \\

\multicolumn{1}{r|}{Refined} & Es wird erwartet, dass die Kostüme nach Angaben des NRF {\color{red}1,2 Milliarden Dollar} aus den {\color{blue}6,9 Milliarden Dollar} \\
\multicolumn{1}{r|}{} & ausmachen. \\

\\

\multicolumn{2}{l}{Example 3} \\
\toprule
\multicolumn{1}{r|}{Source} & It was with this piece of Bedouin wisdom that the first ever chairman Wolfgang Henne described the history and\\
\multicolumn{1}{r|}{} & fascination behind the ``Helping Hands'' society .\\

\multicolumn{1}{r|}{Reference} & Mit dieser Beduinenweisheit beschrieb der erste Vorsitzende Wolfgang Henne die Geschichte und Faszination des\\
\multicolumn{1}{r|}{} & Vereins ``Helfende Hände''. \\

\multicolumn{1}{r|}{Original} & {\color{red}Der erste Vorsitzende Wolfgang Henne} beschrieb {\color{blue}mit dieser erste Weisheit in Bedouin"} die Geschichte und\\
\multicolumn{1}{r|}{} & Faszination hinter der ``Helenden Hands'' Gesellschaft \\

\multicolumn{1}{r|}{Refined} & {\color{blue}Mit diesem Stück Bedouin-Weisheit} beschrieb {\color{red}der erste Vorsitzende Wolfgang Henne} jemals die Geschichte und\\
\multicolumn{1}{r|}{} & Faszination hinter der ``Heling Hands'' Gesellschaft\\

\end{tabular}
\caption{
Sample translations on {\wmtende}. We show the translation from a latent variable sampled from the prior (\emph{Original}) and the translation after one refinement step in the continuous space with the learned score function (\emph{Refined}).
We emphasize phrases whose positions are swapped in the refinement process in red and blue.
}
\label{tab:gen}
\end{table*}

\section{Related Work}
\label{related}

\paragraph{Learning}
Our training objective is closely related to the score matching objective~\cite{hyvarinen05estimation}, with the following differences.
First, we approximate the gradient of the data log density using a proxy gradient, whereas this term is replaced by the Hessian of the energy in the original score matching objective.
Second, we only consider samples from the prior.
\citet{saremi18deep} proposed a denoising interpretation of the Parzen score objective~\citep{vincent11connection} that avoids estimating the Hessian.
Although score function estimation that bypasses energy estimation was found to be unstable~\citep{alain14what,saremi18deep}, it has been successfully applied to generative modeling of images~\citep{song19generative}.

\paragraph{Inference}
While we categorize inference methods for machine translation as (1) discrete search, (2) hybrid optimization~\citep{shu20latent} and (3) continuous optimization (this work) in Section~\ref{sec:background}, another line of work relaxes discrete search into continuous optimization~\citep{hoang17towards,gu18neural,tu20engine}.
By using Gumbel-softmax relaxation~\citep{maddison17concrete,jang17categorical}, they train an inference network to generate target tokens that maximize the log probability under a pretrained model.

\paragraph{Gradient-based Inference}
Performing gradient descent over structured outputs was mentioned in ~\citet{lecun06tutorial}, and has been successfully applied to many structured prediction tasks~\citep{belanger16structured,wang16proximal,belanger17end}.
Other work performed gradient descent over the latent variables to optimize objectives for a wide variety of tasks, including chemical design~\citep{gomez18automatic} and text generation~\citep{mueller17sequence}


\paragraph{Generation by Refinement}
Refinement has a long history in text generation. 
The retrieve-and-refine framework retrieves an (input, output) pair from the training set that is similar to the test example, and performs edit operations on the corresponding output~\citep{sumita91experiments,song16two,hashimoto2018retrieve,weston18retrieve,gu18search}.
The idea of refinement has also been applied in automatic post-editing~\citep{novak16iterative,grangier17quickedit}.

\section{Conclusion}

We propose an efficient inference procedure for non-autoregressive machine translation that refines translations purely in the continuous space.
Given a latent variable model for machine translation, we train an inference network to approximate the gradient of the marginal log probability with respect to the target sentence, using only the latent variable.
This allows us to use gradient based optimization to find a target sentence at inference time that approximately maximizes the marginal log probability.
As we avoid discrete search over a large vocabulary, our inference procedure is more efficient than previous inference procedures that refine in the token space.

We compare our approach with a recently proposed delta inference procedure that optimizes jointly in discrete and continuous space on three machine translation datasets: {\wmtende}, {\wmtroen} and {\iwsltdeen}.
With the same underlying latent variable model, the proposed inference procedure using a learned score function has following advantages:
(1) it is twice as fast as delta inference, and (2) it is able to find target sentences resulting in higher marginal probabilities and BLEU scores.

While we showed that iterative inference with a learned score function is effective for spherical Gaussian priors, more work is required to investigate if such an approach will also be successful for more sophisticated priors, such as Gaussian mixtures or normalizing flows.
This will be particularly interesting, as recent study showed latent variable models with a flexible prior give high test log-likelihoods, but suffer from poor generation quality as inference is challenging~\citep{lee20discrepancy}.

\section*{Acknowledgments}

This work was supported by Samsung Advanced Institute of Technology (Next Generation Deep Learning: from pattern recognition to AI), Samsung Research (Improving Deep Learning using Latent Structure) and NSF Award 1922658 NRT-HDR: FUTURE Foundations, Translation, and Responsibility for Data Science. KC thanks CIFAR, eBay, Naver and NVIDIA for their support.

\bibliography{anthology,emnlp2020}

\begin{thebibliography}{43}
\expandafter\ifx\csname natexlab\endcsname\relax\def\natexlab#1{#1}\fi

\bibitem[{Alain and Bengio(2014)}]{alain14what}
Guillaume Alain and Yoshua Bengio. 2014.
\newblock What regularized auto-encoders learn from the data-generating
  distribution.
\newblock \emph{J. Mach. Learn. Res.}, 15(1):3563--3593.

\bibitem[{Belanger and McCallum(2016)}]{belanger16structured}
David Belanger and Andrew McCallum. 2016.
\newblock Structured prediction energy networks.
\newblock In \emph{Proceedings of the 33nd International Conference on Machine
  Learning}, volume~48 of \emph{{JMLR} Workshop and Conference Proceedings},
  pages 983--992.

\bibitem[{Belanger et~al.(2017)Belanger, Yang, and McCallum}]{belanger17end}
David Belanger, Bishan Yang, and Andrew McCallum. 2017.
\newblock End-to-end learning for structured prediction energy networks.
\newblock In \emph{Proceedings of the 34th International Conference on Machine
  Learning}, volume~70 of \emph{Proceedings of Machine Learning Research},
  pages 429--439. {PMLR}.

\bibitem[{Chan et~al.(2019)Chan, Kitaev, Guu, Stern, and
  Uszkoreit}]{chan19kermit}
William Chan, Nikita Kitaev, Kelvin Guu, Mitchell Stern, and Jakob Uszkoreit.
  2019.
\newblock {KERMIT:} generative insertion-based modeling for sequences.
\newblock \emph{arXiv preprint arxiv:1906.01604}.

\bibitem[{Ghazvininejad et~al.(2019)Ghazvininejad, Levy, Liu, and
  Zettlemoyer}]{gha19mask}
Marjan Ghazvininejad, Omer Levy, Yinhan Liu, and Luke Zettlemoyer. 2019.
\newblock Mask-predict: Parallel decoding of conditional masked language
  models.
\newblock In \emph{Proceedings of the 2019 Conference on Empirical Methods in
  Natural Language Processing and the 9th International Joint Conference on
  Natural Language Processing, {EMNLP-IJCNLP}}, pages 6111--6120.

\bibitem[{G{\'o}mez-Bombarelli et~al.(2018)G{\'o}mez-Bombarelli, Wei, Duvenaud,
  Hern{\'a}ndez-Lobato, S{\'a}nchez-Lengeling, Sheberla, Aguilera-Iparraguirre,
  Hirzel, Adams, and Aspuru-Guzik}]{gomez18automatic}
Rafael G{\'o}mez-Bombarelli, Jennifer~N. Wei, David Duvenaud, Jos{\'e}~Miguel
  Hern{\'a}ndez-Lobato, Benjam{\'\i}n S{\'a}nchez-Lengeling, Dennis Sheberla,
  Jorge Aguilera-Iparraguirre, Timothy~D. Hirzel, Ryan~P. Adams, and Al{\'a}n
  Aspuru-Guzik. 2018.
\newblock Automatic chemical design using a data-driven continuous
  representation of molecules.
\newblock \emph{ACS Central Science}, 4(2):268--276.

\bibitem[{Grangier and Auli(2017)}]{grangier17quickedit}
David Grangier and Michael Auli. 2017.
\newblock Quickedit: Editing text \& translations via simple delete actions.
\newblock \emph{arXiv preprint arXiv:1711.04805}.

\bibitem[{Gu et~al.(2018{\natexlab{a}})Gu, Bradbury, Xiong, Li, and
  Socher}]{gu18non}
Jiatao Gu, James Bradbury, Caiming Xiong, Victor O.~K. Li, and Richard Socher.
  2018{\natexlab{a}}.
\newblock Non-autoregressive neural machine translation.
\newblock In \emph{6th International Conference on Learning Representations,
  {ICLR}}.

\bibitem[{Gu et~al.(2018{\natexlab{b}})Gu, Im, and Li}]{gu18neural}
Jiatao Gu, Daniel~Jiwoong Im, and Victor O.~K. Li. 2018{\natexlab{b}}.
\newblock Neural machine translation with gumbel-greedy decoding.
\newblock In \emph{Proceedings of the Thirty-Second {AAAI} Conference on
  Artificial Intelligence}, pages 5125--5132. {AAAI} Press.

\bibitem[{Gu et~al.(2019)Gu, Liu, and Cho}]{gu19insertion}
Jiatao Gu, Qi~Liu, and Kyunghyun Cho. 2019.
\newblock Insertion-based decoding with automatically inferred generation
  order.
\newblock \emph{Trans. Assoc. Comput. Linguistics}, 7:661--676.

\bibitem[{Gu et~al.(2018{\natexlab{c}})Gu, Wang, Cho, and Li}]{gu18search}
Jiatao Gu, Yong Wang, Kyunghyun Cho, and Victor O.~K. Li. 2018{\natexlab{c}}.
\newblock Search engine guided neural machine translation.
\newblock In \emph{Proceedings of the Thirty-Second {AAAI} Conference on
  Artificial Intelligence}, pages 5133--5140. {AAAI} Press.

\bibitem[{Hashimoto et~al.(2018)Hashimoto, Guu, Oren, and
  Liang}]{hashimoto2018retrieve}
T.~Hashimoto, K.~Guu, Y.~Oren, and P.~Liang. 2018.
\newblock A retrieve-and-edit framework for predicting structured outputs.
\newblock In \emph{Advances in Neural Information Processing Systems
  (NeurIPS)}.

\bibitem[{Hoang et~al.(2017)Hoang, Haffari, and Cohn}]{hoang17towards}
Cong Duy~Vu Hoang, Gholamreza Haffari, and Trevor Cohn. 2017.
\newblock Towards decoding as continuous optimisation in neural machine
  translation.
\newblock In \emph{Proceedings of the 2017 Conference on Empirical Methods in
  Natural Language Processing}, pages 146--156. Association for Computational
  Linguistics.

\bibitem[{Hyv{\"{a}}rinen(2005)}]{hyvarinen05estimation}
Aapo Hyv{\"{a}}rinen. 2005.
\newblock Estimation of non-normalized statistical models by score matching.
\newblock \emph{J. Mach. Learn. Res.}, 6:695--709.

\bibitem[{Jang et~al.(2017)Jang, Gu, and Poole}]{jang17categorical}
Eric Jang, Shixiang Gu, and Ben Poole. 2017.
\newblock Categorical reparameterization with gumbel-softmax.
\newblock In \emph{5th International Conference on Learning Representations}.
  OpenReview.net.

\bibitem[{Kaiser et~al.(2018)Kaiser, Bengio, Roy, Vaswani, Parmar, Uszkoreit,
  and Shazeer}]{kaiser18fast}
Lukasz Kaiser, Samy Bengio, Aurko Roy, Ashish Vaswani, Niki Parmar, Jakob
  Uszkoreit, and Noam Shazeer. 2018.
\newblock Fast decoding in sequence models using discrete latent variables.
\newblock In \emph{Proceedings of the 35th International Conference on Machine
  Learning, {ICML}}, pages 2395--2404.

\bibitem[{Kim and Rush(2016)}]{kim16sequence}
Yoon Kim and Alexander~M. Rush. 2016.
\newblock Sequence-level knowledge distillation.
\newblock In \emph{Proceedings of the 2016 Conference on Empirical Methods in
  Natural Language Processing, {EMNLP}}, pages 1317--1327.

\bibitem[{Kingma and Ba(2015)}]{kingma15adam}
Diederik~P. Kingma and Jimmy Ba. 2015.
\newblock Adam: {A} method for stochastic optimization.
\newblock In \emph{3rd International Conference on Learning Representations,
  {ICLR}}.

\bibitem[{Kingma and Welling(2014)}]{kingma14auto}
Diederik~P. Kingma and Max Welling. 2014.
\newblock Auto-encoding variational bayes.
\newblock In \emph{2nd International Conference on Learning Representations,
  {ICLR} 2014, Banff, AB, Canada, April 14-16, 2014, Conference Track
  Proceedings}.

\bibitem[{Kudo and Richardson(2018)}]{kudo18sentencepiece}
Taku Kudo and John Richardson. 2018.
\newblock Sentencepiece: {A} simple and language independent subword tokenizer
  and detokenizer for neural text processing.
\newblock In \emph{Proceedings of the 2018 Conference on Empirical Methods in
  Natural Language Processing}, pages 66--71. Association for Computational
  Linguistics.

\bibitem[{LeCun et~al.(2006)LeCun, Chopra, Hadsell, Huang, and
  et~al.}]{lecun06tutorial}
Yann LeCun, Sumit Chopra, Raia Hadsell, Fu~Jie Huang, and et~al. 2006.
\newblock A tutorial on energy-based learning.
\newblock In \emph{Predicting Structured Data}. MIT Press.

\bibitem[{Lee et~al.(2018)Lee, Mansimov, and Cho}]{lee18deterministic}
Jason Lee, Elman Mansimov, and Kyunghyun Cho. 2018.
\newblock Deterministic non-autoregressive neural sequence modeling by
  iterative refinement.
\newblock In \emph{Proceedings of the 2018 Conference on Empirical Methods in
  Natural Language Processing}, pages 1173--1182.

\bibitem[{Lee et~al.(2020)Lee, Tran, Firat, and Cho}]{lee20discrepancy}
Jason Lee, Dustin Tran, Orhan Firat, and Kyunghyun Cho. 2020.
\newblock On the discrepancy between density estimation and sequence
  generation.
\newblock \emph{arXiv preprint arxiv:2002.07233}.

\bibitem[{Ma et~al.(2019)Ma, Zhou, Li, Neubig, and Hovy}]{ma19flowseq}
Xuezhe Ma, Chunting Zhou, Xian Li, Graham Neubig, and Eduard~H. Hovy. 2019.
\newblock Flowseq: Non-autoregressive conditional sequence generation with
  generative flow.
\newblock In \emph{Proceedings of the 2019 Conference on Empirical Methods in
  Natural Language Processing}, pages 4281--4291.

\bibitem[{Maddison et~al.(2017)Maddison, Mnih, and Teh}]{maddison17concrete}
Chris~J. Maddison, Andriy Mnih, and Yee~Whye Teh. 2017.
\newblock The concrete distribution: {A} continuous relaxation of discrete
  random variables.
\newblock In \emph{5th International Conference on Learning Representations}.

\bibitem[{Mansimov et~al.(2019)Mansimov, Wang, and Cho}]{mansimov19generalized}
Elman Mansimov, Alex Wang, and Kyunghyun Cho. 2019.
\newblock A generalized framework of sequence generation with application to
  undirected sequence models.
\newblock \emph{arXiv preprint arxiv:1905.12790}.

\bibitem[{Mueller et~al.(2017)Mueller, Gifford, and
  Jaakkola}]{mueller17sequence}
Jonas Mueller, David~K. Gifford, and Tommi~S. Jaakkola. 2017.
\newblock Sequence to better sequence: Continuous revision of combinatorial
  structures.
\newblock In \emph{Proceedings of the 34th International Conference on Machine
  Learning}, volume~70 of \emph{Proceedings of Machine Learning Research},
  pages 2536--2544. {PMLR}.

\bibitem[{Novak et~al.(2016)Novak, Auli, and Grangier}]{novak16iterative}
Roman Novak, Michael Auli, and David Grangier. 2016.
\newblock Iterative refinement for machine translation.
\newblock \emph{arXiv preprint arXiv:1610.06602}.

\bibitem[{Ott et~al.(2019)Ott, Edunov, Baevski, Fan, Gross, Ng, Grangier, and
  Auli}]{ott19fairseq}
Myle Ott, Sergey Edunov, Alexei Baevski, Angela Fan, Sam Gross, Nathan Ng,
  David Grangier, and Michael Auli. 2019.
\newblock fairseq: {A} fast, extensible toolkit for sequence modeling.
\newblock In \emph{Proceedings of the 2019 Conference of the North American
  Chapter of the Association for Computational Linguistics: Human Language
  Technologies}, pages 48--53. Association for Computational Linguistics.

\bibitem[{Post(2018)}]{post18call}
Matt Post. 2018.
\newblock A call for clarity in reporting {BLEU} scores.
\newblock In \emph{Proceedings of the Third Conference on Machine Translation:
  Research Papers}, pages 186--191. Association for Computational Linguistics.

\bibitem[{Saremi et~al.(2018)Saremi, Mehrjou, Sch{\"{o}}lkopf, and
  Hyv{\"{a}}rinen}]{saremi18deep}
Saeed Saremi, Arash Mehrjou, Bernhard Sch{\"{o}}lkopf, and Aapo
  Hyv{\"{a}}rinen. 2018.
\newblock Deep energy estimator networks.
\newblock \emph{arXiv preprint arxiv:1805.08306}.

\bibitem[{Sennrich et~al.(2016)Sennrich, Haddow, and
  Birch}]{sennrich16edinburgh}
Rico Sennrich, Barry Haddow, and Alexandra Birch. 2016.
\newblock Edinburgh neural machine translation systems for {WMT} 16.
\newblock In \emph{Proceedings of the First Conference on Machine Translation,
  {WMT}}, pages 371--376.

\bibitem[{Shu et~al.(2020)Shu, Lee, Nakayama, and Cho}]{shu20latent}
Raphael Shu, Jason Lee, Hideki Nakayama, and Kyunghyun Cho. 2020.
\newblock Latent-variable non-autoregressive neural machine translation with
  deterministic inference using a delta posterior.
\newblock In \emph{Proceedings of the Thirty-Fourth AAAI Conference on
  Artificial Intelligence}.

\bibitem[{Song and Ermon(2019)}]{song19generative}
Yang Song and Stefano Ermon. 2019.
\newblock Generative modeling by estimating gradients of the data distribution.
\newblock In \emph{Advances in Neural Information Processing Systems 32: Annual
  Conference on Neural Information Processing Systems 2019}, pages
  11895--11907.

\bibitem[{Song et~al.(2016)Song, Yan, Li, Zhao, and Zhang}]{song16two}
Yiping Song, Rui Yan, Xiang Li, Dongyan Zhao, and Ming Zhang. 2016.
\newblock Two are better than one: An ensemble of retrieval- and
  generation-based dialog systems.
\newblock \emph{arXiv preprint arxiv:1610.07149}.

\bibitem[{Sumita and Iida(1991)}]{sumita91experiments}
Eiichiro Sumita and Hitoshi Iida. 1991.
\newblock Experiments and prospects of example-based machine translation.
\newblock In \emph{29th Annual Meeting of the Association for Computational
  Linguistics}, pages 185--192. Association for Computational Linguistics.

\bibitem[{Teh et~al.(2003)Teh, Welling, Osindero, and Hinton}]{teh03energy}
Yee~Whye Teh, Max Welling, Simon Osindero, and Geoffrey~E. Hinton. 2003.
\newblock Energy-based models for sparse overcomplete representations.
\newblock \emph{J. Mach. Learn. Res.}, 4:1235--1260.

\bibitem[{Tu et~al.(2020)Tu, Pang, Wiseman, and Gimpel}]{tu20engine}
Lifu Tu, Richard~Yuanzhe Pang, Sam Wiseman, and Kevin Gimpel. 2020.
\newblock Engine: Energy-based inference networks for non-autoregressive
  machine translation.

\bibitem[{Vaswani et~al.(2017)Vaswani, Shazeer, Parmar, Uszkoreit, Jones,
  Gomez, Kaiser, and Polosukhin}]{vaswani17attention}
Ashish Vaswani, Noam Shazeer, Niki Parmar, Jakob Uszkoreit, Llion Jones,
  Aidan~N. Gomez, Lukasz Kaiser, and Illia Polosukhin. 2017.
\newblock Attention is all you need.
\newblock In \emph{Advances in Neural Information Processing Systems 30: Annual
  Conference on Neural Information Processing Systems}, pages 5998--6008.

\bibitem[{Vincent(2011)}]{vincent11connection}
Pascal Vincent. 2011.
\newblock A connection between score matching and denoising autoencoders.
\newblock \emph{Neural Computation}, 23(7):1661--1674.

\bibitem[{Wainwright and Jordan(2008)}]{wainwright08graphical}
Martin~J. Wainwright and Michael~I. Jordan. 2008.
\newblock Graphical models, exponential families, and variational inference.
\newblock \emph{Foundations and Trends in Machine Learning}, 1(1-2):1--305.

\bibitem[{Wang et~al.(2016)Wang, Fidler, and Urtasun}]{wang16proximal}
Shenlong Wang, Sanja Fidler, and Raquel Urtasun. 2016.
\newblock Proximal deep structured models.
\newblock In \emph{Advances in Neural Information Processing Systems 29: Annual
  Conference on Neural Information Processing Systems}, pages 865--873.

\bibitem[{Weston et~al.(2018)Weston, Dinan, and Miller}]{weston18retrieve}
Jason Weston, Emily Dinan, and Alexander~H. Miller. 2018.
\newblock Retrieve and refine: Improved sequence generation models for
  dialogue.
\newblock In \emph{Proceedings of the 2nd International Workshop on
  Search-Oriented Conversational AI, SCAI@EMNLP 2018, Brussels, Belgium,
  October 31, 2018}, pages 87--92. Association for Computational Linguistics.

\end{thebibliography}
\bibliographystyle{acl_natbib}

\appendix

\section{Training objective}\label{sec:app}
\label{app:objective}

\begin{align}
\notag
\min_\psi\: &\Big\Vert \Big( - E_\psi(\zfin) + E_\psi(\zini)\Big) -\Big(\tau(\zfin) - \tau(\zini)\Big) \Big\Vert^2 \\
\notag
\approx \min_\psi\: &\Big\Vert \Big((\zfin - \zini)^\intercal \cdot \gradz (- E_\psi(\zini)) \Big) \\
\label{eq:ebm2}
& -\Big( (\zfin - \zini)^\intercal \cdot \gradz \tau(\zini) \Big) \Big\Vert^2 \\
\notag
\approx \min_\psi\: &\Big\Vert \big(\zfin - \zini\big)^\intercal \Big(\gradz (- E_\psi(\zini)) - \gradz \tau(\zini) \Big) \Big\Vert^2 \\
\notag
\approx \min_\psi\: &\Big(\gradz (-E_\psi(\zini)) - \gradz \tau(\zini) \Big)^\intercal \big(\zfin - \zini\big) \\
\notag
&\big(\zfin - \zini\big)^\intercal \Big(\gradz (-E_\psi(\zini)) - \gradz \tau(\zini) \Big) \\
\notag
\approx \min_\psi\: &\Big(\gradz (-E_\psi(\zini)) - \gradz \tau(\zini) \Big)^\intercal \big\Vert\zfin - \zini\big\Vert^2 \\
\label{eq:ebm5}
&\Big(\gradz (-E_\psi(\zini)) - \gradz \tau(\zini) \Big) \\
\notag
\approx \min_\psi\: &\Big\Vert\gradz (-E_\psi(\zini)) - \gradz \tau(\zini) \Big\Vert^2 \\
\notag
\approx \min_\psi\: &\Big\Vert \gradz(-E_\psi(\z)) \Big\Vert^2 + \Big\Vert \gradz\tauz \Big\Vert^2 \\
\label{eq:ebm7}
& - 2\,\Big( \gradz(-E_\psi(\z))^\intercal \cdot \gradz\tauz \Big) \\
\notag
\approx \min_\psi\: &\Big\Vert \gradz E_\psi(\z) \Big\Vert^2 + 2\,\Big( \gradz E_\psi(\z)^\intercal \cdot \gradz\tauz \Big)
\end{align}

\noindent Eq.~\ref{eq:ebm2} follows from linear approximation, as 
\begin{align*}
-\big(E_\psi(\zfin) - E_\psi(\zini)\big) &\approx (\zfin - \zini)^{\intercal} \cdot \gradz (- E_\psi(\zini)) \\
\tau(\zfin) - \tau(\zini) &\approx (\zfin - \zini)^{\intercal} \cdot \gradz \tau(\zini)
\end{align*}

\noindent $\Vert\zfin - \zini\Vert^2$ in Eq.~\ref{eq:ebm5} can be eliminated, as dividing the objective with a positive constant does not change the solution. The second term in Eq.~\ref{eq:ebm7} is also a constant with respect to $\psi$, hence can be ignored. 

\end{document}